\documentclass[times, 10pt,twocolumn]{article} 
\usepackage{latex8}
\usepackage{times}
\usepackage{graphicx}
\usepackage{authblk}
\usepackage{multirow}  
\usepackage{url}

\usepackage{hyperref}
\begin{document}  

\title{\fontsize{14}{12}\selectfont Foreground-Background Imbalance Problem in Deep Object Detectors: A Review}
\author[1]{Joya Chen}  
\author[2]{Qi Wu} 
\author[1]{Dong Liu} 
\author[1]{Tong Xu} 
\affil[1]{University of Science and Technology of China, Hefei 230027, China}
\affil[2]{Institute of Intelligent Machines, Chinese Academy of Sciences}
\affil[ ]{\small\texttt{\{chenjoya, wqdfj\}@mail.ustc.edu.cn \quad \{dongeliu, tongxu\}@ustc.edu.cn}}
\maketitle
\thispagestyle{empty}

\begin{abstract}
Recent years have witnessed the remarkable developments made by deep learning techniques for object detection, a fundamentally challenging problem of computer vision. Nevertheless, there are still difficulties in training accurate deep object detectors, one of which is owing to the foreground-background imbalance problem. In this paper, we survey the recent advances about the solutions to the imbalance problem. First, we analyze the characteristics of the imbalance problem in different kinds of deep detectors, including one-stage and two-stage ones. Second, we divide the existing solutions into two categories: sampling heuristics and non-sampling schemes, and review them in detail. Third, we experimentally compare the performance of some state-of-the-art solutions on the COCO benchmark. Promising directions for future work are also discussed.       
\footnotetext{This work was supported by the National Key Research and Development Program of China under Grant 2018YFA0701603, and by the Natural Science Foundation of China under Grants 61772483 and 61931014. \emph{(Corresponding author: Dong Liu.)}}
\end{abstract} 

\Section{Introduction}
Object detection, which consists of recognizing the categories and identifying the locations of object instances that appeared in an image, attracts numerous research efforts for several decades. As a fundamental task in computer vision, it is the basis for solving more complex and high-level vision tasks, e.g. instance segmentation~\cite{mask_rcnn}, image caption~\cite{bottom_up_and_top_down}, and scene understanding~\cite{graph_rcnn}. Moreover, it plays a key role in a series of real-world applications, such as autonomous driving, robotic vision, and video surveillance.

In earlier years, the sliding window paradigm with hand-crafted features~\cite{dpm,jones} was widely used for detecting objects. With the rapid development of deep learning techniques~\cite{dl}, deep object detectors~\cite{cascade_rcnn,mask_rcnn,densebox,foveabox,cornernet,tridentnet,fpn,focal_loss,ssd,yolo,yolov2,yolov3,faster_rcnn,fcos,reppoints,centernet,extremenet} quickly come to dominate the research of object detection, and have substantially pushed the detection accuracy forward. Despite the apparent differences in various detection architectures, e.g. one-stage~\cite{ssd,yolo} versus two-stage~\cite{faster_rcnn}, previous works~\cite{ghm,focal_loss,ssd,faster_rcnn,ohem} reveal that the \textit{foreground-background imbalance} problem universally exists in training object detectors, i.e. there is extreme inequality between the number of foreground examples and the number of background examples. Strong evidence~\cite{ghm,focal_loss,ssd,ohem} has shown that the imbalance problem impedes detectors from achieving a higher detection accuracy.

In this paper, we thoroughly review the recent advances in solving the foreground-background imbalance problem. Firstly, as the imbalance problem incurs different consequences among various object detectors, we carefully analyze the characteristics of the imbalance for different object detectors, i.e., anchor-based one-stage, anchor-free one-stage, and two-stage approaches. Subsequently, we divide the solutions into two groups: sampling heuristics~\cite{pisa,ghm,focal_loss,libra_rcnn,faster_rcnn,ohem} and non-sampling schemes~\cite{resobj,sampling_free,ap_loss,dr_loss}, and systematically review the existing solutions for the imbalance problem in detail. Meanwhile, a comparison of their performance is given.  Finally, several promising directions are discussed to inspire future research.

\SubSection{Scope}
As a longstanding difficulty, the class imbalance problem has been studied for a long while in machine learning research.  While the foreground-background imbalance problem in deep object detectors could also be viewed as a class imbalance problem, it is attributed to the large searching space of detectors, rather than the usual causes such as data distribution (i.e., due to biased dataset). Therefore, we only discuss the foreground-background imbalance in object detection.  Furthermore, as the state-of-the-art performance is often achieved by deep object detectors, we will ignore the imbalance solutions in classic non-deep object detectors.

\SubSection{Comparison with Previous Reviews}
Several surveys e.g.~\cite{generic_od} have comprehensively review the object detection tasks, datasets, metrics, and methods. However, they did not specifically discuss the imbalance problems of object detection in detail. Oksuz et al.~\cite{imbalance_review} provide a review for different kinds of imbalance problems in object detection, including class imbalance, scale imbalance, spatial imbalance, and objective imbalance. They did not focus on the imbalance between foregrounds and backgrounds in deep object detectors. We pay attention to the foreground-background imbalance problem and provide a more dedicated review of the solutions to this problem.

\SubSection{Paper Organization}
This paper is organized as follows: Section~\ref{sec2} introduces the research background of deep object detectors with the explanation of the foreground-background imbalance problem. Section~\ref{sec3} describes the solutions to the foreground-background imbalance problem in detail and compares the performance between different solutions. Section~\ref{sec4} concludes the paper and discusses several promising directions.

\Section{Research Background} \label{sec2}
Here we briefly introduce the deep object detectors and the foreground-background imbalance problem. Following the previous work of~\cite{generic_od}, we summarize various deep object detectors into \textit{one-stage} and \textit{two-stage} approaches, but further divide one-stage approach into \textit{anchor-based} one-stage detectors and \textit{anchor-free} one-stage detectors, as the foreground-background imbalance problem has different characteristics for them. For each category, we will analyze what causes the foreground-background imbalance.

\SubSection{Imbalance in One-Stage Object Detectors}

\noindent\textbf{Anchor-Based One-Stage Detectors.} 
With the dense, predefined bounding-boxes (i.e., anchors~\cite{faster_rcnn}) tiled over an image, anchor-based one-stage detectors~\cite{ghm,focal_loss,ssd,yolov2,yolov3,refinedet} could directly recognize objects by refining the locations and classifying the category of these anchors. Early representatives include SSD~\cite{ssd} and YOLOv2~\cite{yolov2} that manage to predict objects on multiple feature levels, which achieve impressive speed/accuracy trade-off. 

However, there is a large gap between the foreground examples and the background example (e.g. $\sim$100 vs. $\sim$100$k$) during training, i.e., foreground-background imbalance. As illustrated in previous works~\cite{ghm,focal_loss,ssd}, this imbalance would impede anchor-based one-stage detectors from becoming more accurate. RetinaNet~\cite{focal_loss}, RefineDet~\cite{refinedet}, and GHM~\cite{ghm} explore different solutions for addressing the imbalance, yielding much better detection accuracy.

\noindent\textbf{Anchor-Free One-Stage Detectors.}
As anchors would introduce multiple hyper-parameters to determine the shape (e.g. scales, aspect ratios), some researchers have started to explore an anchor-free paradigm. Early efforts include DenseBox~\cite{densebox}, YOLO~\cite{yolo}, and CornerNet~\cite{cornernet}, which rely on the central region, the fixed cell, and the key point to determine the initial location, respectively. Their successors could be divided into two categories: center-based~\cite{foveabox,fcos,centernet} and point-based~\cite{reppoints,extremenet} frameworks. Some two-stage approaches also draw lessons from anchor-free one-stage pipelines, e.g. GA-RPN~\cite{garpn}.

In practice, both the key points and the central regions of objects only occupy a small part of the image, while the majority in the image is the background. Despite that anchor-free approaches discard dense anchors to cover objects by key-points/the central regions, they still suffer from the imbalance caused by the overwhelming number of background points or regions, which could be identified as a foreground-background imbalance problem. Therefore, it is not strange that most anchor-free ones apply Focal Loss~\cite{focal_loss} or its variants to address the foreground-background imbalance.

\SubSection{Imbalance in Two-Stage Object Detectors}
To date, two-stage (region-based) object detectors lead the top accuracy on several benchmarks~\cite{pascal_voc,coco}, which shows superiority over one-stage ones in terms of the detection accuracy. These approaches are mainly based on the architecture of Faster R-CNN~\cite{faster_rcnn}, which firstly generates a sparse set of candidate object proposals by a RPN~\cite{faster_rcnn}, then determine the accurate bounding boxes and the classes by convolutional networks. A large number of R-CNN variations~\cite{cascade_rcnn,mask_rcnn,tridentnet,fpn,libra_rcnn,garpn} appear over the years, yielding a large improvement in detection accuracy.

Similar to one-stage detectors, the imbalance problem also exists in two-stage detectors. Firstly, the proposal stage could be viewed as an anchor-based one-stage detector for binary classification (i.e., foreground or background), thus the RPN usually suffers from an extreme imbalance, which requires to apply mini-batch sampling heuristics to alleviate the imbalance. After RPN filters vast background examples, the remaining examples still contain a large number of background examples (e.g., the foreground-to-background ratio is $\sim1:10$). Therefore, the per-region stage is also equipped with mini-batch sampling heuristics.

\begin{figure*} 
    \centering
    \includegraphics[width=0.9\linewidth]{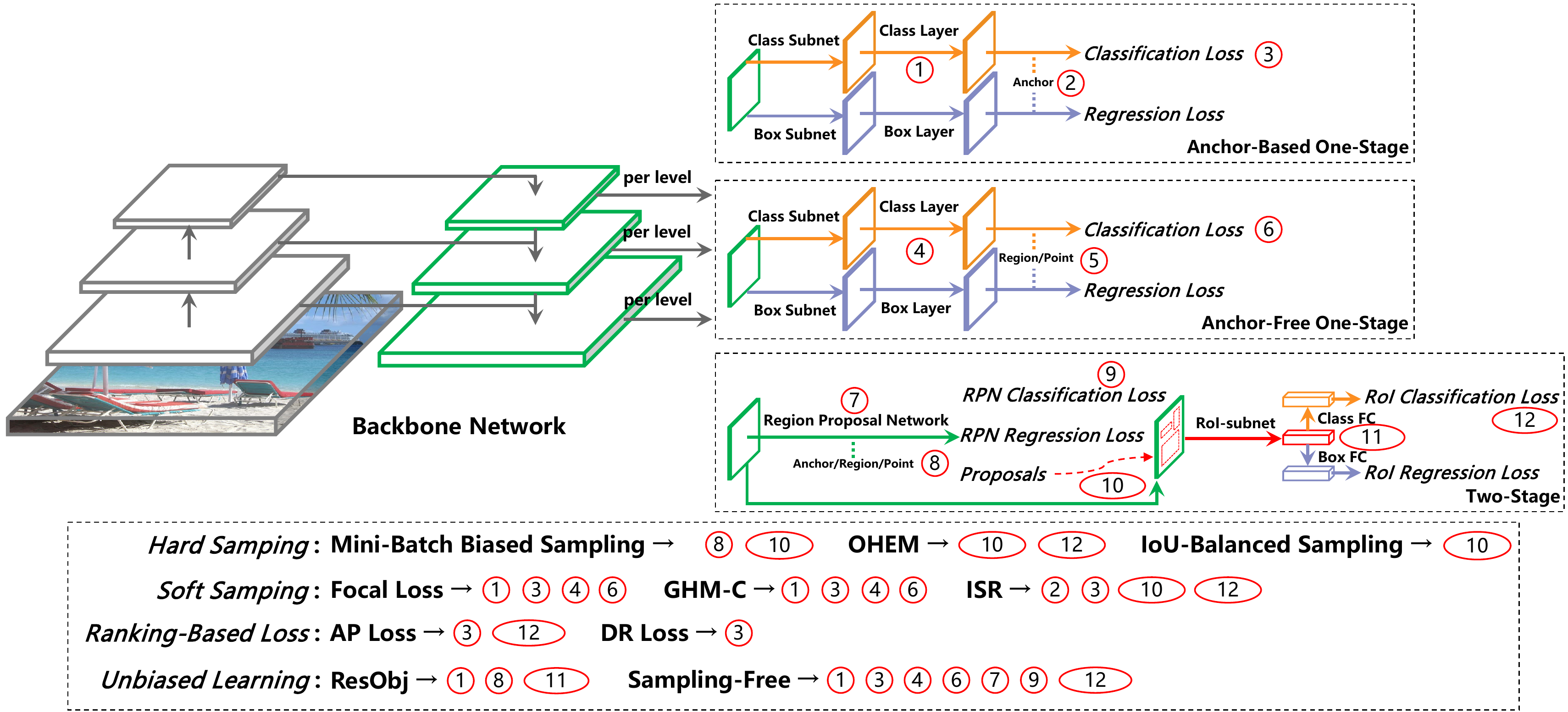}
    \caption{This figure concludes different solutions for addressing the foreground-background imbalance problem in various object detection frameworks (i.e., anchor-based one-stage, anchor-free one-stage, two-stage approaches). These solutions include mini-batch biased sampling~\cite{faster_rcnn}, OHEM~\cite{ohem}, IoU-balanced sampling~\cite{libra_rcnn}, Focal Loss~\cite{focal_loss}, GHM-C~\cite{ghm}, ISA~\cite{pisa}, ResObj~\cite{resobj}, sampling-free~\cite{sampling_free}, AP Loss~\cite{ap_loss}, DR Loss~\cite{dr_loss}. We visualize their scope of the usage in the detection pipeline.}\label{figure1}
\end{figure*}

\Section{Solutions} \label{sec3}
We divide the solutions for addressing the foreground-background imbalance problem into two groups: (1) sampling heuristics, including hard~\cite{libra_rcnn,faster_rcnn,ohem} and soft sampling heuristics~\cite{pisa,ghm,focal_loss}; (2) non-sampling schemes, consisting of ranking-based loss functions~\cite{ap_loss,dr_loss} and unbiased learning mechanisms~\cite{resobj,sampling_free}. In this section, we will describe these solutions in detail.
 
\SubSection{Sampling Heuristics}
In essence, sampling heuristics addresses the imbalance by changing the contribution of each example e.g. put more focuses on rare foreground examples:

\begin{equation}
    L_i^{cls} = \Omega(G, E, P, i) \cdot CE(p_i, g_i), \label{loss_cls}
\end{equation}

\noindent where $CE()$ indicates the cross-entropy cost. $L_i^{cls}$, $p_i$ and $g_i$ denotes the classification loss, predicted probability and the ground-truth label of $i$-th example, respectively. We denote $\Omega()$ as the sampling heuristics, i.e., given the input ground-truths $G$, examples $E$, predictions $P$, and example index $i$, $\Omega(G, E, P, i)$ outputs the weighting factor for $i$-th example. For simplicity, we denote $N_X$ as the number of $X$. Based on Figure~\ref{figure1} and Equation~\ref{loss_cls}, we describe different sampling heuristics as follows.

\subsubsection{Hard Sampling} 

Hard sampling selects a part of training examples while ignores others, i.e., $\Omega(G, E, P, i) \in \{0, 1\}$. We further denote $Pr(G, E, P, i)$ as the probability of $\Omega(G, E, P, i) = 1.$ 

\noindent\textbf{Mini-Batch Biased Sampling.}
It is widely used in two-stage approaches~\cite{faster_rcnn,fpn}, which randomly selects examples by a predefined foreground-to-background ratio $fg/bg$ and the required number of examples $N_Q$. For example, a default setting in Faster R-CNN~\cite{faster_rcnn,fpn} is, randomly sampling 256 anchors, 512 RoIs with a ratio of $fg/bg=1/1$, $fg/bg=1/3$ in RPN, RoI-subnetwork, respectively. For example, $Pr(G, E, P, i)_{g_i>0} = (N_Q/N_{E_{g_i>0}}) \cdot [fg/(fg+bg)]$ is the sampled probability of foreground examples.

\noindent\textbf{OHEM.}
Online hard example mining (OHEM~\cite{ohem}) prefers to sample harder examples than easier ones. A typical form is, $Pr(G, E, P, i) = 1$ only if $CE(p_i, g_i)$ within top 512 values, otherwise $Pr(G, E, P, i) = 0$. Compared with the mini-batch biased sampling, OHEM needs additional memory and causes more training time. 

\noindent\textbf{IoU-Balanced Sampling.}
IoU-balanced sampling~\cite{libra_rcnn} assumes that backgrounds with higher IoU to ground-truths tend to be the harder ones. It evenly splits the IoU interval (e.g. $[0, 0.5)$) into $K$ bins, then randomly selects examples in these bins. Therefore, IoU-balanced sampling saves the extra loss computation required by OHEM. Note that this is only applied for backgrounds, e.g. in $k$-th bin, $Pr(G, E, P, i)_{g_i=0}^k = (N_{Q_{g_i=0}}/N_{E_{g_i=0}}) \cdot (1 / N_{E_{g_i=0}^k})$.

\subsubsection{Soft Sampling}
Soft sampling scales the contribution $\Omega(G, E, P, i)$ of each example during training. Unlike hard sampling, no example is discarded in this way.

\noindent\textbf{Focal Loss.}
The well-known Focal Loss~\cite{focal_loss} modifies the standard cross-entropy loss to dynamically down-weight the contribution of easy examples:
\begin{equation}
    \Omega(G, E, P, i) = \alpha(1- |g_i - p_i|)^\gamma, \label{focal_loss}
\end{equation}
where $\alpha$ and $\gamma$ are the introduced hyper-parameters. Furthermore, it introduces a biased initialization method in the classification layer (See Figure~\ref{figure1}). As reported in the original paper, the RetinaNet~\cite{focal_loss} achieves the optimal accuracy on COCO~\cite{coco} when $\alpha=0.25$ and $\gamma=2$. 

\noindent\textbf{GHM-C.}
Gradient Harmonizing Mechanism (GHM)~\cite{ghm} claims that the imbalance of examples with different attributes (hard/easy and pos/neg) can be implied by the distribution of gradient norm. It also considers that the density of examples with very large gradient norm (very hard examples) as outliers. Therefore, the classification part of GHM (i.e., GHM-C) focuses on the harmony of gradient contribution, i.e., $\Omega(G, E, P, i) = N_E / GD(|g_i - p_i|)$, where $GD(g_i - p_i)$ is the gradient density of $i$-th example. Note that it follows the biased initialization used in Focal Loss.

\noindent\textbf{ISR.}
Importance-based Sample Reweighting (ISR) is the classification part of PISA~\cite{pisa}. It proposes to weight an example according to the evaluation metric (i.e., mAP~\cite{pascal_voc,coco}). In other words, a part of examples is more important than others in an mAP metric, thus ISR puts more focuses on them. To find the important examples, ISR is built on HLR (IoU-Hierarchical Local Rank), which could select examples that have a large impact on the mAP.

\SubSection{Non-Sampling Schemes}
More recently, some studies propose several non-sampling schemes, which discard sampling heuristic to maintain the distribution of examples. We divide them into two groups: ranking-based loss functions~\cite{ap_loss,dr_loss} and unbiased learning mechanisms~\cite{resobj,sampling_free}.

\subsubsection{Ranking-Based Loss Functions}
\noindent\textbf{AP Loss.}
Instead of modifying the classification loss, AP Loss~\cite{ap_loss} proposes to replace the classification task with a ranking task to optimize average precision (AP). It proposes a novel error-driven learning algorithm to effectively optimize the non-differentiable AP based objective function. More specifically, some extra transformations are added to the output score of an one-stage detector to obtain the AP Loss, which includes a linear transformation that transforms the scores to pairwise differences, and a non-linear and non-differentiable ``activation function'' that transform the pair-wise differences to primary terms of AP Loss. Then the AP Loss can be obtained by the dot product between the primary terms and the label vector.

\noindent\textbf{DR Loss.}
Similar to AP Loss~\cite{ap_loss}, distributional ranking (DR) Loss~\cite{dr_loss} manages to address the imbalance by viewing the classification task as a ranking task. It introduces the DR loss to rank the constrained distribution of foreground above that of background candidates. By reweighting the candidates to derive the distribution corresponding to the worst-case loss, the loss can focus on the decision boundary between foreground and background distribution. Besides, DR Loss ranks the expectation of distributions instead of original examples, which reduces the number of pairs in ranking and improves efficiency.

\subsubsection{Unbiased Learning Mechanisms}

\noindent\textbf{ResObj.}
The complicated, heuristic sampling methods could be replaced by a learning-based algorithm. Residual Objectness (ResObj~\cite{resobj}) is a fully learning-based algorithm, which utilizes multiple cascaded objectness-related modules to address the imbalance. ResObj first transfers the imbalance to the objectness module, to down-weight the contributions of overwhelming backgrounds. Subsequently, by building residual connections between objectness-related modules, they reformulate the objectness estimation to a consecutive refinement procedure, thereby progressively addressing the imbalance.

\noindent\textbf{Sampling-Free.}
Sampling-Free mechanism~\cite{sampling_free} is proposed as an alternative to sampling heuristics, which manages to maintain the training stability from initialization. It demonstrates that sampling heuristics for different types of object detectors could be discarded, while similar or better accuracy could be achieved. The authors of~\cite{sampling_free} observed that, unlike common class imbalance that is introduced by data distribution, the foreground-background imbalance should be attributed to the large searching space of detectors, which means it equally exists in training and inference with the same distribution. But sampling heuristics will change this distribution during training, thereby resulting in a mismatch between training and inference. Under their observation, the obstacle that impedes detectors without sampling from yielding high accuracy should be attributed to the instability during training. Motivated by this, they develop a sampling-free mechanism~\cite{sampling_free}, consisting of three schemes: (1) the optimal bias initialization scheme enables the training to be fastly converged under the imbalance; (2) the guided loss scheme avoids the classification loss to be dominated by numerous background examples; (3) the class-adaptive threshold scheme mitigates the confidence shifting problem incurred by the imbalance.

\begin{table*}[h]
    \scriptsize  
    \centering 
    \begin{tabular}{|c|c|c|c|cccc|}
    \hline
    Solutions & Abbreviation & Detector (ResNet-50-FPN) & Codebase & AP & $\Delta$AP & Parameters & Speed \\
    \hline 
    \multirow{4}*{Hard Sampling} & \multirow{2}*{Biased Sampling~\cite{faster_rcnn}} & \multirow{2}*{Faster R-CNN~\cite{faster_rcnn}} & \textsf{maskrcnn-benchmark} & 36.8 & - & 2 & - \\
    \cline{4-8}
    ~ & ~ & ~ & \textsf{mmdetection} & 36.4 & - & 2 & - \\ 
    \cline{2-8}
    ~ & OHEM~\cite{ohem} & Faster R-CNN & \textsf{mmdetection} & 36.6 & +0.2 & 2 & $\downarrow$ \\
    \cline{2-8}
    ~ & IoU-balanced sampling~\cite{libra_rcnn} & Faster R-CNN & \textsf{mmdetection} & 36.8 & +0.4 & 3 & $\downarrow$ \\
    \hline
    \multirow{6}*{Soft Sampling} & \multirow{4}*{Focal Loss~\cite{focal_loss}} & \multirow{4}*{RetinaNet} & \textsf{maskrcnn-benchmark} & 36.3 & - & 2 & - \\
    \cline{4-8}
    ~ & ~ & ~ & \textsf{mmdetection} & 35.6 & - & 2 & - \\
    \cline{4-8}
    ~ & ~ & ~ & \multirow{2}*{\textsf{detectron}} & 33.9 (600$\times$) & - & 2 & - \\
    ~ & ~ & ~ & ~ & 35.7 (800$\times$) & - & 2 & - \\
    \cline{2-8}
    ~ & GHM-C~\cite{ghm} & RetinaNet & \textsf{mmdetection} & 35.8 & +0.2 & 1 & $\downarrow$ \\
    \cline{2-8}
    ~ & ISR~\cite{pisa} & Faster R-CNN & \textsf{mmdetection} & 37.9 & +1.5 & 4 & $\downarrow$ \\ 
    \hline
    \multirow{2}*{Ranking-Based Loss} & AP Loss~\cite{ap_loss} & RetinaNet & \textsf{detectron} & 35.0 & +1.1 & 2 & $\downarrow$  \\
    \cline{2-8}
    ~ & DR Loss~\cite{dr_loss} & Faster R-CNN & \textsf{detectron} & 37.2 & +1.5 & 4 & $\downarrow$ \\
    \cline{1-8}
    \multirow{3}*{Unbiased Learning} & ResObj~\cite{resobj} &  RetinaNet & \textsf{detectron} & 35.4 & +1.3 & More conv & $\downarrow$ \\
    \cline{2-8}
    ~ & \multirow{2}*{Sampling-Free~\cite{sampling_free}} &  RetinaNet & \textsf{maskrcnn-benchmark} & 36.6 & +0.3 & \textbf{0} & $\uparrow$ \\
    \cline{3-8}
    ~ & ~  & Faster R-CNN & \textsf{maskrcnn-benchmark} & \textbf{38.4} & \textbf{+1.6} & \textbf{0} & $\downarrow$ \\
    \hline
    \end{tabular} 
    \caption{This table illustrates the comparison of different solutions for addressing the foreground-background imbalance problem. We report the accuracy (AP), relative accuracy improvement ($\Delta$AP), the number of hyper-parameters (Parameters), and efficiency (Speed) to compare them. These results come from the well-known detection codebases (i.e., \textsf{maskrcnn-benchmark}~\cite{maskrcnn_benchmark}, \textsf{mmdetection}~\cite{mmdetection}, \textsf{detectron}~\cite{detectron}). ``-'' indicates the baseline models. If not specified, we report the performance achieved with backbone ResNet-50-FPN~\cite{resnet,fpn}, input scale $1333\times800$, evaluated on COCO \texttt{minival}~\cite{coco}. We observed that sampling-free~\cite{sampling_free} achieves the highest AP and $\Delta$AP without introduced hyper-parameters. ISR~\cite{pisa} and DR Loss~\cite{ap_loss} also achieve impressively relative accuracy improvement. }\label{table1}
\end{table*}

\subsection{Comparison}

To better review the solutions, we comprehensively compare their performances in Table~\ref{table1}. These results are either from the ablation study of the original paper or from the model description page (i.e., \textsf{MODEL\_ZOO.md}) of the public codebases. For a fair comparison, we select RetinaNet~\cite{focal_loss} and Faster R-CNN~\cite{faster_rcnn} as the baseline models, which apply mini-batch biased sampling and Focal Loss to address the imbalance, respectively. We compare these solutions from four aspects: accuracy, relative accuracy improvement, required hyper-parameters and training speed (compared to without the solution).

\noindent\textbf{Accuracy.}
As presented in Table~\ref{table1}, Sampling-Free and ISR achieve the highest and the second-highest accuracy, respectively. Meanwhile, they keep this advantage in the ``$\Delta$AP'', which refers to the relative accuracy improvement.

\noindent\textbf{Hyper-Parameters.} 
See the ``Parameters'' column in Table~\ref{table1}. Despite ISR and DR Loss obtain the second-highest ``$\Delta$AP'', they introduce the most hyper-parameters (4 hyper-parameters). In contrast, sampling-free introduces no hyper-parameter, whereas other solutions introduce at least 1 hyper-parameter.

\noindent\textbf{Training Speed.}
As the reported training speed may be evaluated on different devices, the speed comparison may not be fair.  Instead, we compare the speed relative to the baseline models. The ``Speed'' column in Table~\ref{table1} shows that the majority of solutions are slower than baseline models, except for RetinaNet with the sampling-free mechanism.

\Section{Discussions} \label{sec4}

In this paper, we present a comprehensive review of the foreground-background imbalance in deep object detectors. Our review first analyzes different deep object detectors and explain the causes of the foreground-background imbalance. Subsequently, we categorize and describe the existing solutions for addressing the imbalance, including sampling heuristics and non-sampling schemes. Finally, we experimentally compare the performances of different solutions. Based on these summarizations, we discuss two important questions about the foreground-background imbalance:

First, what is the cause of the foreground-background imbalance problem? It appears to be the large searching space of deep object detectors. To date, both anchor-based and anchor-free frameworks follow a dense prediction paradigm, which produces numerous background examples during training. This imbalance may disappear if a detector with a small searching space would be developed.

Second, which solution should be used? As shown in Table~\ref{table1}, Sampling-Free~\cite{sampling_free} has achieved the highest relative accuracy improvement in two-stage approaches, without introduced hyper-parameters. DR Loss~\cite{dr_loss} achieves the highest relative accuracy improvement in one-stage approaches. Despite more hyper-parameters introduced in DR Loss, its success suggests that a solution should be developed according to an evaluation metric, as mAP is a ranking-related metric. If there are multiple evaluation metrics in the real-world application, we recommend the sampling-free mechanism as a baseline method, whose code can be found at \href{https://github.com/ChenJoya/sampling-free}{https://github.com/ChenJoya/sampling-free}.

\bibliographystyle{latex8}
\bibliography{ai.bib}

\end{document}